\documentclass{article}

\usepackage{arxiv}

\usepackage[utf8]{inputenc} % allow utf-8 input
\usepackage[T1]{fontenc}    % use 8-bit T1 fonts
\usepackage{hyperref}       % hyperlinks
\usepackage{url}            % simple URL typesetting
\usepackage{booktabs}       % professional-quality tables
\usepackage{amsfonts}       % blackboard math symbols
\usepackage{nicefrac}       % compact symbols for 1/2, etc.
\usepackage{microtype}      % microtypography
\usepackage{lipsum}		% Can be removed after putting your text content
\usepackage{graphicx}
\usepackage{doi}

\date{}

\title{The Poison of Alignment}

\author{
        Aibek Bekbayev\thanks{Equal contribution\\Correspondence to: Sungbae Chun <sungbae@goat.ai>}\\
        GOAT AI\\
        \And
        Sungbae Chun\footnotemark[1]\\
        GOAT AI\\
        \And
        Yerzat Dulat\footnotemark[1]\\
        Higgsfield AI\\
        \And
        James Yamazaki\footnotemark[1]\\
        GOAT AI\\
}

\begin{document}

\maketitle

\begin{abstract}
From the perspective of content safety issues, alignment has shown to limit large language models' (LLMs) harmful content generation. This intentional method of reinforcing models to not respond to certain user inputs seem to be present in many modern open-source instruction tuning datasets such as OpenAssistant or Guanaco. We introduce a novel insight to an instruction-tuned model's performance affected by the presence of alignment in supervised fine-tuning dataset. To be specific, we noticed that alignment acts as if it is poisoning the instruction dataset. Experimentally, we demonstrate that aligned answers significantly worsen the performance of the resulting fine-tuned model's on various reasoning benchmarks such as Big Bench (BBH), Massive Multitask Language Understanding (MMLU), Human Eval, and Discrete Reasoning Over Paragraphs (DROP), performing worse than the counterpart tuned without alignment by 4-33\%.

%Current research trend is mainly focused on knowledge distillation of large foundation models. Our research team sought a way to enhance our LM's performance via novel data processing mechanisms on our app's huge user-bot interaction dataset. Our model learns from the mixture of signals from ChatGPT, Text-da-Vinci, GPT-3, and GPT-4 produced during user-bot interaction. \{evaluation and results here\}
\end{abstract}

\section{Introduction}
Emerging power of Large Language Models (LLMs) has shown impressive ability to perform greatly on complex benchmarks, such as Human Eval~\cite{chen2021codex} and Big Bench (BBH)~\cite{srivastava2022beyond}, and in professional examination settings such as SAT, GRE, and LSAT with few or no examples~\cite{gpt3}. Despite LLMs not yet reaching peak human performance in professional exams or complex benchmarks, the performance gap between LLMs and top-scoring humans has steadily narrowed in recent years with the help of scaling and better data processing techniques~\cite{scalinglaws}.

Particular attention in the recent literature was drawn to knowledge distillation models, including Vicuna\cite{chiang2023vicuna}, Alpaca\cite{alpaca}, and the more recent Orca\cite{mukherjee2023orca}, that claims performances comparable to that of ChatGPT~\cite{chatgpt}. For instance, Mukherje et al.\cite{mukherjee2023orca} reported that Orca surpassed ChatGPT on the Vicuna evaluation set, using GPT-4~\cite{openai2023gpt4} for assessment, and achieved parity with ChatGPT on most evaluation tasks in their study.

Despite the spike in both research and open-source community, a recent study by Gudibande et al.\cite{gudibande2023false} suggests that known distillation models mainly emulate the style and "learn" dialogue format, rather than unleash reasoning capabilities or factual accuracy. The study found that while models fine-tuned on ChatGPT responses generate well-structured output resembling the original model, the content often contained errors or deviated significantly from the topic.

Our study complements the study by Gudibande et al.~\cite{gudibande2023false}, as we observe substantial improvements on reasoning benchmarks such as Massive Multitask Language Understanding (MMLU) ~\cite{hendryckstest2021, hendrycks2021ethics} or BBH following supervised fine-tuning (SFT) with finely grained datasets. 
% These improvements are documented in leaderboards like LMSys\cite{zheng2023judging} and the Huggingface open-llm leaderboard\cite{open-llm-leaderboard, eval-harness,clark2018think,zellers2019hellaswag,lin2022truthfulqa, hendryckstest2021}. 
Our experiments consistently demonstrated better performance in reasoning skills over the base model, with the smaller models showing the most noticeable improvement.

In this paper, we present novel insights into dataset cleaning methods for SFT: alignment as the source of instruction dataset poisoning. Our dataset, collected from our GoatChat app, substantially improves the fine-tuned model's performance over the base model in MMLU and BBH. This empirically augments the findings of Gudibande et al.~\cite{gudibande2023false}. We consistently observe significant improvements in benchmarks such as MMLU, BBH, Discrete Reasoning Over Paragraphs (DROP)~\cite{dua-etal-2019-drop}, and Human Eval~\cite{chen2021codex} at scale using the amount of data comparable or less to one of open-source fine-tuning datasets. All models in this paper are evaluated using InstructEval~\cite{chia2023instructeval}, with the exception of proprietary models.

\section{Background}
\textbf{Data cleaning.} The analysis of dataset cleaning methods~\cite{lee2021deduplicating, wenzek2019ccnet} has made notable strides in recent years, significantly enhancing the performance of LLMs trained on public datasets such as C4~\cite{raffel2020c4} and The Pile~\cite{gao2020pile}. The importance of dataset cleaning was firmly investigated in the study by the Falcon team~\cite{penedo2023refinedweb} in which authors have implemented various methods of dataset cleaning, including custom processing pipeline for CC-like datasets and fuzzy/exact deduplication. Results have shown that dataset cleaning takes a vital part in performance of LLMs. Recent paper by Zhou et al. ~\cite{zhou2023lima} that focuses on importance of data for supervised instruction fine-tuning claims that data quality is of greater importance rather than data quantity.

A comprehensive study by Penedo et al. (the Falcon team)~\cite{penedo2023refinedweb} evaluated the impacts of various data filtering methods on the performance of the resulting models. Their study shows that their experiments, conducted on both small-scale (1B and 3B parameters trained on 27GT and 60GT, respectively) and full-scale (1B parameters trained on 350GT), revealed that cleaned web-crawl datasets can serve as viable training datasets boosting overall performance of LLMs. This finding challenges the prevailing belief that curated datasets generally outperform web-crawled datasets in LLMs. Furthermore, the study also showed that deduplicating The Pile led to performance benefits for models trained on it. This emphasizes the need for cleaning and deduplicating data to achieve optimal model performance, even when working with pre-curated datasets like The Pile. These observations reinforce a key principle in model training: the quality of the data is crucial. This aligns with the conclusion of the work of Zhou et al.~\cite{zhou2023lima}  that the quality of data has a greater impact on model performance than data quantity.

\textbf{Supervised fine-tuning.} After InstructGPT~\cite{chatgpt} was introduced by OpenAI team, there have been numerous studies that conduct SFT on an open-source LLMs with main trigger being the release of LLaMA~\cite{touvron2023llama} by Meta AI. Many research teams built SFT models on top of LLaMA, and the most prominent ones are Vicuna~\cite{chiang2023vicuna}, Stanford Alpaca~\cite{alpaca}, and Orca~\cite{mukherjee2023orca}. However, this active trend towards SFT faced a criticism as well. The works of Gudibande et al.~\cite{gudibande2023false} indicated that during SFT, the models performance do not increase over the bare LLMs' performance.

\textbf{Data poisoning.} With active development of SFT models, there have been efforts to study exploitability of LLMs upon instruction tuning. The works of Wan et al.~\cite{wan2023poisoning} demonstrated that LLM's behaviour can be manipulated with as few as hundreds of poisonous examples. 
Furthermore, Shu et al.~\cite{shu2023exploitability} discussed more non-straightforward poisoning of SFT dataset.
Inspired by the above studies, it seems possible that aligned answers in SFT datasets may nudge a model's behaviour into an undesirable direction, acting as a poisonous contaminant.

\section{Method}
\subsection{Dataset collection}
\begin{figure}
    \centering
    \includegraphics[width=0.5\linewidth]{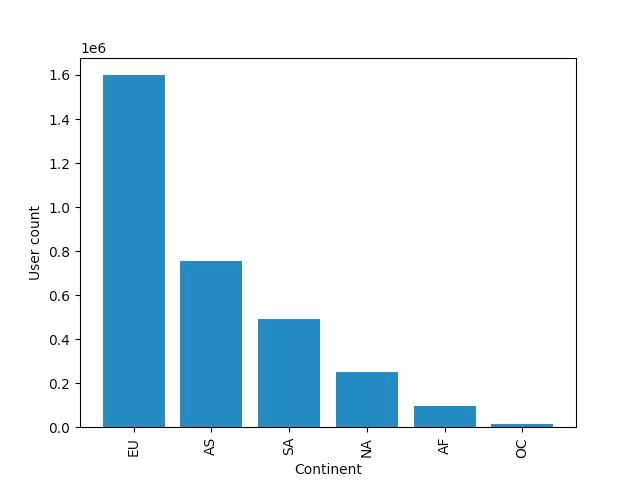}
    \caption{Distribution of users by continents. Continent code (CC) is used: EU - Europe, AS - Asia, SA - South America, NA - North America, AF - Africa, and OC - Oceania}
    \label{fig:users_count}
\end{figure}

For the dataset collection we have utilized our top-rated app GoatChat that has over 3 million users (see Figure \ref{fig:users_count} for detailed user statistics). GoatChat provides a simple interface for interaction with AI assistant. All users sign a terms of agreement to collect their data to be used in the further research.

\subsection{Dataset cleaning and deduplication}
\textbf{Basic quality filtering.} Our private dataset collected from GoatChat was mainly composed of the interaction of user and AI assistant. From the structure of our app, there were several kinds of defects in our dataset that can possibly impose unwanted behaviour thus had to be cleaned. Our first cleaning pipeline was aimed at filtering out the following defective data points: API failures, low-quality chats, and mixed language.

By API failures we mean instances in which one-to-one correspondence between user-bot messages did not hold. There were several reasons why such kind of data heterogeneity happened, such as the case in which user's message was not delivered to API possibly due to aggressive language content in the input chat (racism, sexism, etc.) or the case in which user made two consecutive messages due to bug. It is important to underline that we assume the latter behaviour as "failure" because our app's chat was meant to have strictly alternating chat sequence between the user and the bot.

By low-quality chats we mean data points that were considered to have non-informative content. At the message level, we eliminated data with short input text as it empirically was shown that it rarely contains informative inputs. Additionally, we filtered out whole chat sessions with low number of average tokens per message (due to assumption of non-informativeness) and with numerous repeated queries (spamming). Upon an investigation of the data, we found out that the former contained mostly just nonsense texts or plain numbers.
We call the resulting filtered version of the dataset as \textit{GoatAssistant}.

\textbf{Dataset merge}
For our further work, we have merged GoatAssistant dataset with Guanaco~\cite{guanaco} dataset to enhance the diversity of resulting dataset. %write something about guanaco

\textbf{Exact and fuzzy deduplication.}
For exact and fuzzy deduplication we have used the works of Lee et al.~\cite{lee2021deduplicating} and used the thresholds as ones suggested in original study. We have performed deduplication at chat-level and dropped $17.4\%$ of original dataset.

\textbf{Alignment removal.} 
We have noted that the majority of aligned answers do not contain informative responses to the user query, which is evident considering the fact that the model's response is passive, i.e. the model is reluctant to provide the exact information that user requested. A strong model that we are aiming to get at the end should be able respond to a user query as informative as possible, and additionally, alignment often contains input prompts that are not necessarily inappropriate. This filtering removed about a third of our dataset, and because it was our novel method of dataset cleaning, we also performed ablation study to isolate the effect of aligned answers reflected onto the tuned model.

\section{Experimental Setup}
We employed all our computations on one node of 8xA100 NVIDIA GPU. Training was done using bfloat16 and DeepSpeed~\cite{deepspeed} ZeRO-3. All models were initially trained for $3$ consecutive epochs with checkpointing on each half of the epoch. However, we empirically observed that training over $1$ epoch degrades the model quality and reverted to using only $1$ epoch with checkpointing on half of the epoch. For memory optimization, we used x-formers~\cite{xformers} and gradient checkpointing~\cite{gradientcheckpoint}. We kept effective batch size at 512 during training of 7B models. We used standard AdamW~\cite{adamw} optimizer with learning rate of 1e-4 and betas set to (0.9, 0.999) with warmup steps being about $7\%$ of all training steps amount. 

\section{Evaluation}
We evaluate our model on various reasoning benchmarks: MMLU, BBH, HumanEval, and DROP.

\textbf{MMLU} seeks to evaluate LLM proficiency across a vast spectrum of domains, ranging from humanities to hard sciences. It is composed of 15,908 multiple-choice questions sourced from academic examinations, university course materials, and specialized texts. This benchmark is crucial in measuring a model's capacity for comprehensive real-world textual comprehension and its aptitude for extracting knowledge from extensive corpora.

\textbf{BBH} was introduced to characterize emerging capabilities in LLMs and delineate potential limitations. It encompasses 204 tasks, delving into areas such as linguistics, biology, and software development. The benchmark, calibrated against state-of-the-art architectures from dense to sparse transformers, offers invaluable insights into performance trends, scale-associated enhancements, and task-centric challenges.

\textbf{HumanEval} is specifically tailored to assess functional correctness in algorithmic tasks. With 164 hand-crafted programming problems, which include function signatures, docstrings, and unit tests, it tests LLMs on comprehension, reasoning, and algorithmic synthesis. This benchmark provides a unique lens into an LLM's ability to not just replicate but genuinely understand and produce syntactically and semantically accurate code.

Lastly, \textbf{DROP} benchmark propels reading comprehension evaluation by accentuating intricate textual reasoning. This adversarially-generated dataset, with 96k questions, demands nuanced reference resolution coupled with discrete operations such as sorting and arithmetic. It presents a formidable challenge for models, pushing them to transition from basic information retrieval to a more profound, multi-dimensional comprehension.

\begin{table}[h]
    \centering
    \caption{7B model comparison}
    \begin{tabular}{|c|c|c|}
        \hline
         Task       &LLaMA 2      &Our model  \\
         \hline 
         MMLU       &45.94           &49.31 \\
         BBH        &32.04           &35.69\\
         Human Eval &14.02           &12.20\\
         DROP       &31.57         &28.10\\
         \hline
    \end{tabular}
    \label{tab:goat7bscore}
\end{table}

We notice that with our novel data processing method, we achieve a better performance than the underlying foundation model by a significant margin in MMLU and BBH.

\subsection{Ablation study}

For ablation study, we have produced 2 datasets: the first one is our GoatAssitant and Guanaco~\cite{guanaco}, and the second one is the first dataset without alignment. We trained both models under the same training setups specified before.
%probably we need to name datasets for better readability

\begin{table}[h]
    \centering
    \caption{Ablation study results}
    \label{tab:goat7balignmentcomp}
    \begin{tabular}{|c|c|c|}
        \hline
         Task &With alignment &No alignment\\
         \hline 
         MMLU       &45.63  &49.31 (\textbf{8.1\%})\\
         BBH        &34.28  &35.69  (\textbf{4.1\%})\\
         HumanEval  &9.15   &12.20  (\textbf{33.3\%})\\
         DROP       &22.61  &28.10  (\textbf{24.3\%})\\
         \hline
    \end{tabular}
\end{table}

As it can be seen from Table \ref{tab:goat7balignmentcomp}, we see that when the model was trained on our aligned dataset, it did not improve over the base model, which confirms the study by Gudibande et al.~\cite{gudibande2023false}. However, we also observe a remarkable performance increase upon fine-tuning our model on the cleaned version of our dataset.
Therefore, it seems that the negative impact of alignment distorted the performance boost of previous fine-tuning methods, so that the models did not show a significant improvement on reasoning abilities, leading to the underestimation of reasoning ability gain upon SFT.

\section{Limitations}

This study, as it was done based off on LLaMA 2, inherits most of its limitations including data biases, lack of world understanding and the hallucination. Methods suggested in this study may be inapplicable for tailoring the model for certain behaviour and generally oriented only towards research purposes and was tested only in research environments. Concerning the models, one obvious limitation is the lack of computing resources that did not allow us to fully fine-tune models with size over 7B.

\section{Conclusion}

In this study, we propose a new perspective of the instruction tuning that the presence of alignment behaves similar to the dataset poisoning. We demonstrate that alignment at the stage of SFT harms the model's performance by a significant margin (4-33\% in reasoning benchmarks).
Additionally, this study reassures the emerging effectiveness of thorough dataset cleaning and preparation applied to the task of supervised instruction fine-tuning despite the criticism that supervised fine-tuning is mainly a formatting task. Namely, we uncover the details about our dataset that can be of use in understanding of efficient dataset building for supervised instruction fine-tuning as well as describe our thorough data cleaning pipeline.

\section*{Acknowledgments}
This work was supported by GOAT AI. We thank Dos Baha for the organisation and funding of this research project; Zhenisbek Assylbekov for his valuable feedback; Yerbol Kopzhassar for his key role in communicating with externals in securing the necessary hardware; Akzhol Ibraimov, Alexey Muryshkin, and Arman Oralov  for their contribution in data collection.

\bibliographystyle{ieeetr}
\bibliography{references}

\vfill

\end{document}